\documentclass[lettersize, journal,compsoc]{IEEEtran}
\usepackage{amsmath,amsfonts}
\usepackage{algorithmic}
\usepackage{array}
\usepackage[caption=false,font=normalsize,labelfont=sf,textfont=sf]{subfig}
\usepackage{textcomp}
\usepackage{stfloats}
\usepackage{url}
\usepackage{verbatim}
\usepackage{graphicx}
\usepackage{cite}

\usepackage{tcolorbox}
\usepackage[capitalise]{cleveref}
\usepackage{mathtools}
\usepackage{booktabs}
\usepackage{siunitx}
\newcommand{\IGNORE}[1]{}
\DeclarePairedDelimiterX\Set[1]\{\}{
    
    #1
}

\newcommand{\AlgA}{\alpha}
\newcommand{\AlgB}{\beta}
\newcommand{\AlgC}{\gamma}
\newcommand{\AlgD}{\delta}
\newcommand{\AlgE}{\epsilon}

\begin{document}

\title{SIERRA: A Modular Framework for Research Automation and Reproducibility}

\author{John Harwell$^{1}$ and Maria Gini$^{1}$,\IEEEmembership{Fellow,IEEE}
\thanks{$^{1}$University of Minnesota
        {\tt\small \{harwe006,gini\}@umn.edu}}
}

\maketitle

\begin{abstract}
  Modern intelligent systems researchers form hypotheses about system behavior
  and then run experiments using one or more independent variables to test their
  hypotheses.  We present SIERRA, a novel framework structured around that idea
  for accelerating research development and improving reproducibility of
  results. SIERRA accelerates research by automating the process of generating
  executable experiments from queries over independent variables(s), executing
  experiments, and processing the results to generate deliverables such as
  graphs and videos. It shifts the paradigm for testing hypotheses from
  procedural (``Do these steps to answer the query'') to declarative (``Here is
  the query to test--GO!''), reducing the burden on researchers. It employs a
  modular architecture enabling easy customization and extension for the needs
  of individual researchers, thereby eliminating manual configuration and
  processing via throw-away scripts.  SIERRA improves reproducibility of
  research by providing automation independent of the execution environment (HPC
  hardware, real robots, etc.) and targeted platform (arbitrary simulator or
  real robots). This enables exact experiment replication, up to the limit of
  the execution environment and platform, as well as making it easy for
  researchers to test hypotheses in different computational environments.
\end{abstract}


\section{Introduction}

In modern intelligent systems research, the majority of researcher time is spent
on two types of tasks: science and engineering. Science tasks consist of
developing \emph{AI elements}, defined as a new mathematical model, tool, or
algorithm, while engineering tasks consist of configuring and running
experiments for testing the new AI element, and some aspects of processing
results. Frequently, it is only after science tasks have been nearly completed
for a project that researchers consider the crucial issue of reproducibility,
leaving them little time to ensure their work can be replicated by others.  The
difficulties of reproducibility are further compounded by the nature of the
tools used to meet the engineering needs of a project: ad-hoc toolchains and
scripts which are quickly thrown together on a per-project basis, and reused,
modified, or duplicated on the fly. Usually, these toolchains and scripts are
for dealing with ``accidental complexities''~\cite{Afzal2021-reproduce}; that
is, with engineering difficulties unrelated to the challenges of the science
itself. Examples include: handling different configurations for specific
platforms, such as ROS~\cite{ROS2009}, or execution environments, such as
SLURM~\cite{SLURM2003} clusters, or for processing and visualizing experimental
results; e.g., statistically summarizing data and generating graphs. Clearly,
this approach is prone to errors and to reinventions of the wheel between
research groups and individual researchers. In this paper, we present SIERRA, an
open source framework for automating engineering tasks to improve
reproducibility. SIERRA automates the process of hypothesis testing and results
processing, and handles details for platforms, execution environments, data
processing, and results visualization to reduce the burden on researchers and
allowing them to focus on the ``science'' aspects of research: creative
exploration of data, hypothesis testing, and experimental design.

\section{Motivation and related work}\label{sec:motivation-and-rw}
%
Our motivation in presenting SIERRA is based on our understanding of three
pressing needs in the intelligent systems community. First, the need for better
\emph{automation} of the engineering tasks that many researchers
perform. Second, the need for better \emph{reproducibility}, which is one of the
fundamental problems of intelligent systems
research~\cite{SwarmRob2018-reproduce}. While there is some debate on the exact
criteria for reproducibility, there is general agreement that it is not a binary
designation, but a
spectrum~\cite{Gundersen2019-reproduce,Gundersen2022-reproduce}. Some
methodologies and guidelines exist for helping to increasing reproducibility of
the ``science'' parts of
research~\cite{Gundersen2022-reproduce,Gundersen2018-reproduce}, and a few
tools~\cite{CodeOcean2019-reproduce,DataFed2020-reproduce}. However, few tools
exists for improving the reproducibility of the engineering aspects of research,
which is often intertwined with reproducibility of scientific
results~\cite{Gundersen2022-reproduce}; exceptions
include~\cite{WebotsHPC2021,SwarmRob2018-reproduce}.  Third, any tool meeting
the first two criteria must have a \emph{low barrier to adoption}. In other
words, if the provided research automation or reproducibility guarantees are
difficult to integrate with implementations not designed with them in mind, they
will be much less likely to be adopted. Taken together, a tool that meets these
needs will reduce the barriers to collaboration among researchers in similar
areas which take different approaches, or whose labs have otherwise different
implementations of shared ideas.

We begin with the idea of a \emph{research query} in intelligent systems
research: a query of an independent variable over some range.  Examples include:
``How will this algorithm perform in this scenario with this range of inputs?'',
``What are the practical limits of this algorithm?'', and ``How does this
algorithm compare to other similar algorithms?'' Research queries are different
than scientific hypotheses, which are possible explanations for an observed
phenomenon or answers to a posed research query.  Each ``value'' of the
independent variable in this range forms the basis for an
\emph{experiment}. Experiments take a given value of the independent variable
and operationalize it in the context of a \emph{platform} such as a simulator or
a run-time executive by adding necessary configuration so that the query can be
executed on the platform.  The set of experiments operationalizing a research
query into something that can be executed is defined as a \emph{batch
  experiment}. By comparing results across experiments in the batch, changes in
system behavior in response to the different ``values'' can be observed.  Each
experiment contains one or more \emph{experimental runs}, which can be
simulations, training runs, or real robot trials. Experimental runs are executed
on an \emph{execution environment} such as a High Performance Computing (HPC)
cluster, a researcher's laptop, or real robot hardware.

Formalizing the above, consider the common five stage pipeline for experimental
validation of a new AI element that is shown
in~\cref{tab:research-pipeline}. With this terminology and research pipeline, we
can now discuss the main motivations behind SIERRA in more detail.
\begin{table*}[htbp!]
  \centering
  \caption{Common research pipeline in modern intelligent systems research, from
    an engineering perspective.\label{tab:research-pipeline}}
  \begin{tabular*}{\textwidth}{ p{1.5cm} p{16cm} }\toprule
    {Stage} &
    {Description/current practice} \\ \midrule

    \rule{0pt}{3ex}Experiment generation &

    A researcher designs a batch experiment to test an AI element. If there is
    randomness in the experimental inputs or in the AI element itself such as a
    random seed, multiple experimental runs are defined for each per experiment
    so that statistically valid inferences about the behavior of the AI element
    can be made.

    \rule{0pt}{4ex}\textbf{Current practice.} Researchers utilize custom and/or
    throw-away scripts or otherwise manually set parameters defining the
    experiments\cite{Afzal2021-reproduce}. \\

    \rule{0pt}{4ex}Experiment execution &

    The researcher runs the batch experiment, collecting data about different
    aspects of the AI element.

    \rule{0pt}{4ex}\textbf{Current practice.} Researchers use custom scripts to
    configure their chosen execution environment, such as a High Performance
    Computing (HPC) environment, and then run their experiments on it. There is
    often tight coupling between the execution environment and the targeted
    platform in the scripts, making reuse difficult. Further manual
    configuration is required for real robot applications, such as synchronizing
    the experimental inputs and configuration on each robot. \\
    \rule{0pt}{4ex}Experiment results processing &

    The researcher processes the collected data to generate statistical insights
    about the performance of their AI element, its limitations, and its
    strengths.

    \rule{0pt}{4ex}\textbf{Current practice.} Researchers use mature libraries
    for processing experiment data, such as \verb|pandas|. Scripts for analysis
    are frequently written for the specific pipeline instance; that is, they are
    not reusable between the development of one algorithm and another by a given
    researcher or across research groups. \\
    \rule{0pt}{4ex}Deliverable generation &

    The researcher generates visualizations from processed data which can be
    used to inform further refinements and tuning; common visualizations include
    graphs and videos. If the results are satisfactory, then these deliverables
    are polished to make them camera-ready and included on relevant publications
    or technical reports on the AI element.

    \rule{0pt}{4ex}\textbf{Current practice.} Researchers utilize custom and/or
    throwaway scripts to generate graphs from processed data using
    \verb|matplotlib| or other toolkits. \\

    \rule{0pt}{4ex}Deliverable comparison &

    The researcher generates comparative visualizations of their AI elements and
    other similar elements in the field for inclusion in publications or
    technical reports, in order to show how their contribution improves or is
    different from the state of the art.\rule[-1.5ex]{0pt}{2ex}

    \rule{0pt}{4ex}\textbf{Current practice.} Researchers utilize custom and/or
    throwaway scripts to generate comparative visualizations. \\

    \bottomrule
  \end{tabular*}
\end{table*}

\begin{enumerate}

\item

  \emph{Automation.} From~\cref{tab:research-pipeline}, we note the following
  important insight: most stages contain substantial engineering tasks that are
  performed manually by researchers; these ``accidental complexities'' are
  frequently non-trivial, and slow down the actual research. Dealing with such
  complexities takes valuable researcher time even with many toolkits available
  to help, such as \verb|pandas| and \verb|matplotlib|. In robotics research,
  existing automation in simulation only targets parts of the
  pipeline~\cite{WebotsHPC2021}; similarly for real
  robots~\cite{SwarmRob2018-reproduce,Robotarium2016}. Thus, substantial
  researcher time is spent on the ``non-research'' aspects of intelligent
  systems research generally, and robotics research specifically, limiting
  progress and clearly motivating the need for more general-purpose automation.

\item

  \emph{Reproducibility.}  Some challenges to reproducibility in modern
  intelligent systems research include (a) ``dependency hell'', which is the
  problem of reproducing the execution environment to run research software, (b)
  imprecise or missing documentation, which exacerbates (a), (c) code erosion,
  which is the problem of running outdated researcher code in more recent
  execution environments, and (d) a high barrier to integration with existing
  solutions\cite{SwarmRob2018-reproduce}. These issues are non-trivial; recent
  studies found that less than half of academic code from papers at recent AI
  conferences were runnable (not that they reproduced results, but that they ran
  at all), even with the help of the
  authors~\cite{Gundersen2022-reproduce,Bellogin2021-reproduce}.

\item

  \emph{Low barrier to adoption.} Any software tool that addresses the research
  automation and reproducibility needs in modern intelligent systems community
  must be ``low threshold, no ceiling''~\cite{NetLogo2004}. That is, it must
  meet the following criteria. First, it must have an extremely low barrier for
  new users. The barrier will differ across researchers and fields, but some
  desirable characteristics include: (a) minimal, easy to understand
  configuration, (b) ease of reuse of custom configuration and functionality
  across projects and researchers, (c) plug-and-play faculties that do not
  require recompilation or repackaging to incorporate new functionality, and (d)
  high quality documentation and many
  examples~\cite{Gundersen2022-reproduce,Afzal2021-reproduce}. Second, it must
  be designed to be customizable in unknown ways, in order to support rapid
  adoption by researchers in academic and industry labs. In other words, it must
  be able to accommodate the ``unknown unknown'' future needs of researchers.

\end{enumerate}

\subsection{Motivating Use Cases}\label{sec:use-cases}
We present the following two motivating use cases to help ground the broad areas
in which SIERRA is useful in contexts containing characteristics many
researchers will be familiar with in~\cref{fig:uc1,fig:uc2}. 
\begin{figure}[ht]
  \begin{tcolorbox}[title=Use case \# 1: Alice the foraging algorithm designer]

    Alice is a researcher at a university that has developed a new distributed
    task allocation algorithm $\AlgA$ for use in a foraging task where robots
    must coordinate to find objects of interest in an unknown environment and
    bring them to a central location. Alice wants to implement her algorithm so
    she can investigate:
    \begin{itemize}
    \item

      How well it scales with the number of robots, specifically if it remains
      efficient with up to 100 robots in several different scenarios.

    \item
      How robust it is with respect to sensor and actuator noise.

    \item

      How it compares to other similar state of the art algorithms on a foraging
      task: $\AlgB,\AlgC$.

    \end{itemize}

    Alice is faced with the following heterogeneity matrix that she must deal
    with to answer her research queries, \emph{in addition to the technical
      challenges of the AL elements themselves:}\newline

    \begin{tabular}{c c c }
      {Algorithm} &
      {Has randomness?} &
      {Outputs data in?} \\
      \toprule


      $\AlgA$ & Yes & \verb|CSV,rosbag| \\

      \midrule

      $\AlgB$ & Yes & \verb|CSV,rosbag| \\

      \midrule

      $\AlgC$ & No & \verb|CSV,rosbag| \\

      \bottomrule
    \end{tabular}\newline\newline

    Alice is familiar with ROS, and wants to use it with large scale simulated
    and small scale real-robot experiments with TurtleBots. However, for real
    robots she is unsure what data she will ultimately need, and wants to
    capture all ROS messages with \verb|rosbag| to avoid having to redo
    experiments later. She has access to a large SLURM-managed cluster.
  \end{tcolorbox}
  \caption{Motivating use case \#1: robotics research.\label{fig:uc1}}
\end{figure}

\begin{figure}[ht]
  \begin{tcolorbox}[title=Use case \# 2: Alice the contagion modeler]

    Alice has teamed with Bob, a biologist, to model the spread of contagion
    among agents in a population, and how that affects their individual and
    collective abilities to do tasks. She believes her $\AlgA$ algorithm can be
    reused in this context. However, Bob is not convinced and has selected
    several multi-agent models from recent papers: $\AlgD,\AlgE$, and wants
    Alice to compare $\AlgA$ to them. $\AlgD$ was originally developed in
    NetLogo~\cite{NetLogo2004}, for modeling disease transmission in
    animals. $\AlgE$ was originally developed for ARGoS~\cite{ARGoS2012} to
    model the effects of radiation on robots.  All algorithms contain
    randomness.

    Alice is faced with the following heterogeneity matrix that she must deal
    with to answer the research query, \emph{in addition to the technical
      challenges of the AI elements:}\newline

    \begin{tabular}{c c c}
      {Algorithm} &
      {Can run on?} &
      {Input requirements?} \\

      \toprule

      $\AlgA$ & ROS/Gazebo & XML \\

      \midrule

      $\AlgD$ & NetLogo & NetLogo \\

      \midrule

      $\AlgD$ & ARGoS & XML \\

      \bottomrule
    \end{tabular}\newline\newline

    Bob is interested in how the rate of contagion spread varies with agent
    velocity \emph{and} population size. Bob needs to prepare succinct,
    comprehensive visual representations of the results of his research queries
    for a presentation, including visual comparisons of the multi-agent model
    as it runs for each algorithm. He will give Alice a range of parameter
    values to test for each algorithm based on his ecological knowledge, and
    rely on Alice to do the experiments.  Alice does not have access to HPC
    resources, but does have a handful of servers in her lab that she can use.
  \end{tcolorbox}
  \caption{Motivating use case \#2: collaboration in multi-agent
    modeling.\label{fig:uc2}}
\end{figure}

They will be used later in~\cref{sec:pipeline-automation} to concretely
illustrate many features of SIERRA, such as its ability to handle the
heterogeneity matrices in each use case transparently to our imagined
researcher, Alice.
\section{SIERRA Overview}\label{sec:overview}
\begin{figure*}[t]
  \centering
  \includegraphics[width=\linewidth,height=7cm]{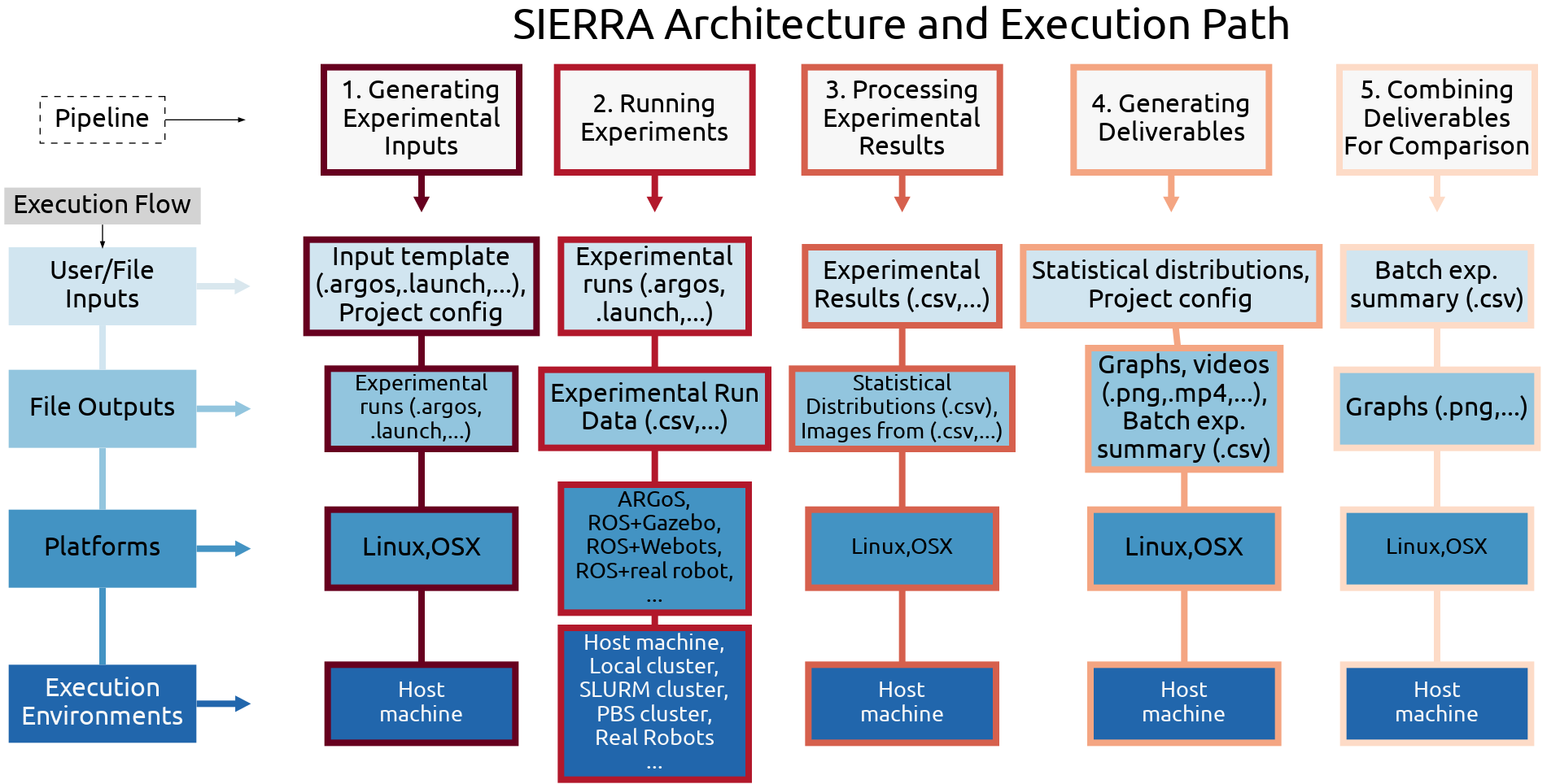}
  \caption{\label{fig:arch} Architecture of SIERRA, organized by pipeline
    stage. Pipeline stages are listed left to right, with an approximate joint
    architectural/functional stack from top to bottom for each stage. ``\dots''
    indicates areas where SIERRA is designed to be configurable or extensible
    with python plugins. ``Host machine'' indicates the machine SIERRA was
    invoked on.}
\end{figure*}

%
In this section, we give a broad overview of SIERRA, a command line tool for
automating the pipeline described in~\cref{tab:research-pipeline}.  High level
details on how SIERRA addresses both issues of reproducibility
and automation in modern intelligent systems research follows below; SIERRA is,
to the best of our knowledge, the first such tool presented in the literature.
An architectural overview is in~\cref{fig:arch}.

\begin{enumerate}
\item

  \emph{Automation.} SIERRA accelerates research cycles by allowing researchers
  to focus on the ``science'' aspects: developing AI elements and designing
  experiments to test them. SIERRA changes the paradigm of the engineering tasks
  researchers must perform from manual and procedural to declarative and
  automated. That is, from ``Do these steps to run the experiment, process the
  data and generate graphs'' to ``Here is the environment and platform, the
  deliverables I want to generate and the data I want to appear on them for my
  research query--GO!''.  Essentially, SIERRA handles the ``backend'' parts of
  research, such as: random seeds, algorithm stochasticity, configuration for a
  given execution environment or platform, generating statistics from
  experimental results, and generating visualizations from processed results.
  By employing declarative specification via command line arguments and YAML
  configuration, it eliminates manual re-configuration of experiments across
  platforms by decoupling the concepts of execution environment and platform
  (see~\cref{tab:support-matrix}); any supported pair can be selected in a
  mix-and-match fashion. Furthermore, it removes the need for throw-away scripts
  for data processing and deliverable generation by providing rich, extensible
  faculties for those pipeline stages.

\item

  \emph{Reproducibility.}  SIERRA supports reusability and reproducibility
  across projects in two main ways. First, through its automation: SIERRA
  experiments are fully reproducible, up to the limit of the platform targeted
  and the execution environment on which the experiments are run. In such cases,
  with a single SIERRA command another researcher Charlie could reproduce
  Alice's exact results and generated deliverables if both were using SIERRA and
  he had access to Alice's code and raw input data.  Second, through its ``low
  threshold, no ceiling'' approach: all aspects of its configuration is done
  through command line switches and YAML configuration, allowing easy reuse
  between projects.
  
  \begin{table}[ht]
  \caption{Current execution environments and supported platforms in
    SIERRA. To the best of our knowledge no automation exists for 
    ARGoS~\cite{ARGoS2012}, Gazebo~\cite{Gazebo2004}, and ROS1~\cite{ROS2009}
     for hypothesis testing and results processing.  The partial
    automation of Webots~\cite{Webots2004}  done in~\cite{WebotsHPC2021}
    is a subset of SIERRA's
    capabilities.\label{tab:support-matrix}}
    
  \begin{tabular}{p{1.5cm} p{4.3cm} p{1.7cm} }
    {Execution\newline{}Environment} &
    {Description} &
    {Supported platforms} \\
    \toprule

    SLURM &

    A SLURM managed HPC cluster. &

    ARGoS, ROS1+Gazebo \\

    \midrule

    Torque/\newline{}MOAB &

    A MOAB managed HPC cluster. &

    ARGoS, ROS1+Gazebo \\

    \midrule

    ADHOC &

    A miscellaneous collection of networked compute nodes a given researcher has
    available. &

    ARGoS, ROS1+Gazebo \\

    \midrule
    Local machine &

    A researcher's local machine to use for small scale testing. &

    ARGoS, ROS1+Gazebo \\

    \midrule

    ROS1+\newline{}TurtleBot3 &

    ROS1 with TurtleBot3~\cite{TurtleBot2020} robots. &

    ROS1+Gazebo, ROS1+robot (real robot) \\

    \bottomrule
  \end{tabular}
 
\end{table}

\item

  \emph{Low barrier to adoption.} SIERRA is designed to have minimal barriers to
  adoption by researchers across disciplines through \emph{in situ} integration
  with existing code implementations, yet also be extensively customizable for
  advanced users; i.e., ``low threshold, no ceiling''. It accomplishes this in
  two ways. First, it is written in the python programming language, which is
  not only ``write once, run anywhere'', but also has a very human readable
  syntax. Second, it is organized into a reusable core and a plugin manager
  which supports any number of plugins of any type that can be used to
  customize nearly every aspect of its implementation of the research pipeline
  shown in~\cref{fig:arch}.  Thus, adding support for a new platform or
  execution environment as simple as implementing a python interface, and
  placing the resulting file(s) on SIERRA's plugin path. Plugins can be written
  in any language; only the bindings must be written in python.  SIERRA is open
  source under the GPLv3 license, allowing researchers to modify it according to
  their needs, and comes with extensive documentation and
  tutorials\footnote{\url{https://swarm-robotics-sierra.readthedocs.io/en/master/}}.

\end{enumerate}

\section{SIERRA pipeline automation}\label{sec:pipeline-automation}
%
SIERRA is designed to automate research queries expressed in a
researcher-defined command line syntax. In SIERRA terminology, this is the
univariate \emph{batch criteria} used to define a batch experiment. SIERRA also
supports \emph{bivariate} batch criteria, in which researchers are interested in
how system behavior changes in response to the values of two independent
variables jointly varying; in such cases, the state space for the batch
experiment is a 2D grid instead of a one dimensional line. SIERRA handles both
types of batch criteria transparently.

For our first use case, these could be:
\begin{itemize}
\item
  \verb|population_size.Log128|, representing univariate experiments with
  $\Set{1,2,4,8,\dots,128}$ agents.
\item
  \verb|ta_policy_set.all.Z100|, representing univariate experiments with one of
  a set of task allocation policies $\Set{\AlgA,\AlgB,\AlgC}$, with the number
  of robots fixed to 100 for all runs.

\item
  \verb|system100 saa_noise.all.C10|, representing bivariate experiments with
  $\Set{1,2,3,\dots,100}$ robots and 10 different levels of noise applied to
  both robot sensors and actuators.
\end{itemize}

For the second use case, this could be:
\noindent \verb|vel.min=1p0.max=10p0.C10 n_agents.Log4096|, representing
bivariate experiments with $\Set{1,2,4,8,\dots,\num{4096}}$ agents and agent
velocities, which will be one of 10 values: $\Set{1.0,2.0,\dots,10.0}$.  The
syntax for expressing research queries is entirely arbitrary, and can be set
according to each researcher's needs; researchers also define parsers for their
syntax.

Once a research query has been operationalized by SIERRA and written to the
filesystem as a batch experiment, SIERRA can execute it by running the
experiment and then process the results; details of the provided automation for
each pipeline stage are shown below. To help further demonstrate SIERRA's
capabilities, we will reference the partial SIERRA commands
in~\cref{fig:uc1:cmd,fig:uc2:cmd} throughout the rest of this section. We note
that running stages $\Set{1,3,3,4,5}$ in sequence is not required; any
topologically ordered subset can be executed. For example, suppose Alice has
just changed the YAML configuration for what deliverables to generate. She could
then instruct SIERRA to run stages $\Set{3,4}$ only by adding
\verb|--pipeline 3 4| to~\cref{fig:uc1:cmd} or~\cref{fig:uc2:cmd}.

\begin{figure}[t]
  \begin{tcolorbox}
\begin{verbatim}
sierra-cli \
--template-input-file=exp.launch \
--platform=platform.ros1gazebo
--project=task_alloc \
--batch-criteria population_size.Log128 \
--controller=task_alloc.alpha \
--robot TurtleBot3 \
--no-master-node\
--exp-setup=exp_setup.T1000.K100
\end{verbatim}
  \end{tcolorbox}
  \caption{ A partial SIERRA command for a batch experiment containing 7
    experiments of some number of runs each; the total \# of Gazebo
    simulations/real robot trials is $7\times \#~runs$. Inputs will be generated
    for ROS1+gazebo and the TurtleBot robot. Experiments will be \num{1000}
    seconds long, with robot controllers running at \SI{100}{\Hz}. SIERRA's
    ability to setup a central ROS node on the ROS master (SIERRA host machine)
    for use in data collection is not needed, and so is disabled. The argument
    to \texttt{--robot} is entirely arbitrary; whatever is passed maps to a set
    of user-defined YAML configuration which allows arbitrary XML
    modifications. This can include specifying ROS nodes to launch, parameters
    to set, etc.\label{fig:uc1:cmd}}
\end{figure}
\begin{figure}[t]
  \begin{tcolorbox}
\begin{verbatim}
sierra-cli \
--template-input-file=exp.launch \
--project=contagion \
--batch-criteria population_size.Log4096 \
--controller=task_alloc.alpha \
--exp-setup=exp_setup.T10000 \
--platform-vc
--n-runs=100
\end{verbatim}
  \end{tcolorbox}
  \caption{ A partial SIERRA command for a batch experiment containing 12
    experiments of \num{100} runs each, for \num{1200} total
    simulations. Experiments will be \num{10000} seconds long, with agent
    controllers running at the default frequency for the
    platform.\texttt{--platform-vc} instructs SIERRA to set up visual capturing
    for the chosen platform: this can be capturing frames to stitch together
    into videos later, or recording videos directly if the platform supports
    it.\label{fig:uc2:cmd}}
\end{figure}

\subsection{Experimental Input Generation}
To generate the batch experiment, researchers provide a template XML file
containing all configuration necessary to answer the research query. Any XML
element SIERRA is not directed to change through a batch criteria or other
plugin will remain unchanged, allowing researchers to set common configuration
options that should remain the same for all experiments and all experimental
runs.

SIERRA currently requires that the template input file be XML, which was chosen
over other input formats for three reasons. First, it is not dependent on
whitespace/tab/spaces for correctness, making it more robust to multiple
platforms, simulators, parsers, users, etc. Second, mature manipulation
libraries exist for python and C++, two of the most common languages in
intelligent systems research, so it should be relatively straightforward for
projects to read experimental definitions from XML. Third, many popular
platforms already support XML input, such as ARGoS, ROS, and WeBots. If a
researcher wants to add support for a platform that does not support XML,
SIERRA's modular architecture makes it easy to do so.

The XML template input file is modified according to the research query, with
one experiment generated for each ``value'' of the independent variable(s). Each
``value'' may correspond to a single change to the template, such as
\verb|population_size.Log32| for changing the number of agents, or it can
correspond to multiple changes, such as \verb|saa_noise.all.C10| for changing
the level of noise applied to multiple sensors and actuators in each
experiment. SIERRA also supports changing additional parts of the template input
file uniquely for each experiment batch, or uniformly for all experiments,
providing unparalleled expressiveness to support research automation through
experiment generation.

SIERRA provides comprehensive support for research that requires multiple
experimental runs in each experiment; this may be required due to randomness in
the agents, e.g., imperfect sensors/actuators on real robots, or algorithm
stochasticity. SIERRA manages this complexity transparently to researchers, and
further provides faculties for supporting idempotency of experiments, up to the
limit of the execution environment and platform. For example, SIERRA can save
generated random seeds, ensuring that if a platform respects the random seed,
then stage 2 of the pipeline in~\cref{tab:support-matrix} is idempotent.

In our first use case, Alice can put any common parameter options in the XML
template file in a \verb|<common>| section and then unique subsections for each
algorithm: \verb|<alpha>|,\verb|<beta>|, etc. She could also give each algorithm
its own XML file and duplicate the \verb|common| section for each, according to
her preference. To handle the stochasticity of $\AlgA,\AlgB$, she can tell
SIERRA to do multiple runs per experiment by adding a \verb|--n-runs| argument
to~\cref{fig:uc1:cmd}. In our second use case, Alice can specify all input
parameters in XML for all algorithms, and then add a new SIERRA platform plugin
for the NetLogo simulator, which transforms the NetLogo template input file into
XML via \verb|xslt| or other tool. Alice can then generate experimental inputs
for each platform from the \emph{same} research query as easily as adding
\verb|--platform.{argos,ros1gazebo,netlogo}| to~\cref{fig:uc2:cmd}.

\subsection{Running Experiments}
After a batch experiment has been operationalized and written to the filesystem,
SIERRA can execute it in full or only arbitrary subsets of the experiments. For
example, if experiments \#10--12 keep crashing, Alice can enable more debugging
in her code and then re-run the problematic experiments by adding
\verb|--exp-range=10:12| to~\cref{fig:uc1:cmd}. We note that SIERRA's automation
for this pipeline stage enables it to provide concurrent execution of
experimental runs for platforms that do not support it natively, such as Gazebo,
and to utilize intrinsic parallelism for platforms that do support it, such as
NetLogo.

Experimental run inputs are executed using GNU parallel~\cite{GNUParallel2011}
on a selected execution environment and targeted platform
(see~\cref{tab:support-matrix} for SIERRA's current support matrix). SIERRA
handles the necessary configuration for all supported platforms and execution
environments, allowing researchers to transparently switch between them with
minimal code changes; cross-compiling or re-architecting may be necessary
depending on the nature of researcher code, and the selected platform.  This
effectively makes the question of ``Where can I run my experiment?'' logistical
and declarative, rather than technical and procedural.

In our first use case, Alice can run her code on her laptop during development
by adding \verb|--exec-env=hpc.local| to~\cref{fig:uc1:cmd}. Then, once she is
confident in her algorithm's correctness, she only has to pass
\verb|--exec-env=hpc.slurm| instead to tell SIERRA to run her code on her SLURM
cluster. She will only have to submit a SLURM job containing her SIERRA
invocation with a given set of resources and SIERRA will figure out everything
else. For her real robot experiments, she will only need to change
\verb|--exec-env=robots.turtlebot3| and \verb|--platform=platform.rosrobot| to
tell SIERRA to run with real TurtleBots. In our second use case, after the
necessary code tweaks, Alice can tell SIERRA to run her simulations on her adhoc
network of compute nodes by adding \verb|--exec-env=adhoc|
to~\cref{fig:uc1:cmd}.

\subsection{Processing Experimental Results}
After a batch experiment has been finished (or even part of it has), SIERRA can
process outputs from arbitrary subsets of experiments in the batch using the
\verb|--exp-range|, analogously to the previous stage. Results processing is
independent of experiment execution; that is, SIERRA's plugin framework
homogenizes reading of experimental results recorded in arbitrary formats. To
process results, researchers need to specify which experimental outputs and
types of statistics they are interested in, and SIERRA will then do one or more
of the following. First, statistical distribution generation across experimental
runs for the selected experiments in the batch (intra-experiment statistics), as
well as across experiments in a batch (inter-experiment
statistics). Inter-experiment statistics are necessary so that summary
performance and behavioral measurements can be generated later; only using the
averaged results from the statistical distribution generation in
intra-experiment statistics is often insufficient. Second, converting output
\verb|CSV| files into heatmap images (see~\cref{fig:stage4-ex-graphs}) that can
be stitched together into videos during stage 4.

In both use cases, Alice can generate statistics for all simulations by telling
SIERRA that her simulation data are  stored in \verb|CSV| format by
adding~\verb|--storage-medium=storage.csv| to~\cref{fig:uc1:cmd}. For her real
robot experiments in the first use case she can write a storage plugin that
converts \verb|rosbag| files into \verb|pandas| dataframes, and generate
equivalent statistics. In either case, if she initially specified 95\%
confidence intervals on a set of line graphs, she could switch to box and
whisker plots by changing \verb|--dist-stats=conf95| to \verb|--dist-stats=bw|.

\subsection{Generating Deliverables}
\begin{figure}[t]
  \centering
    \includegraphics[width=0.49\linewidth]{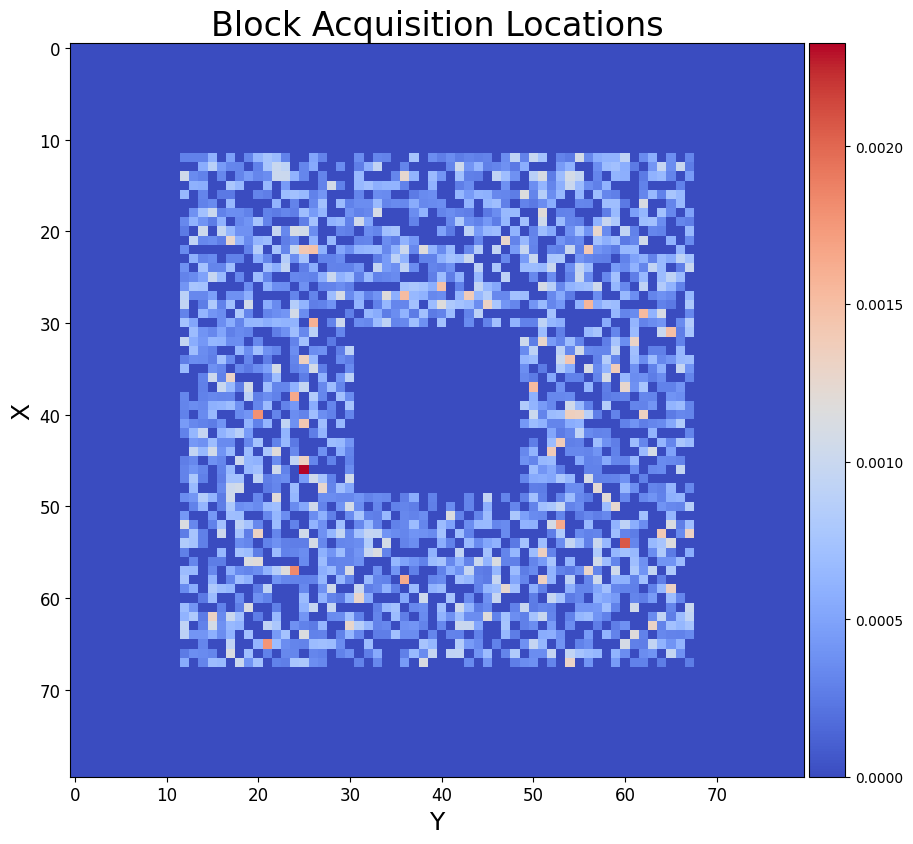}
    \includegraphics[width=0.49\linewidth]{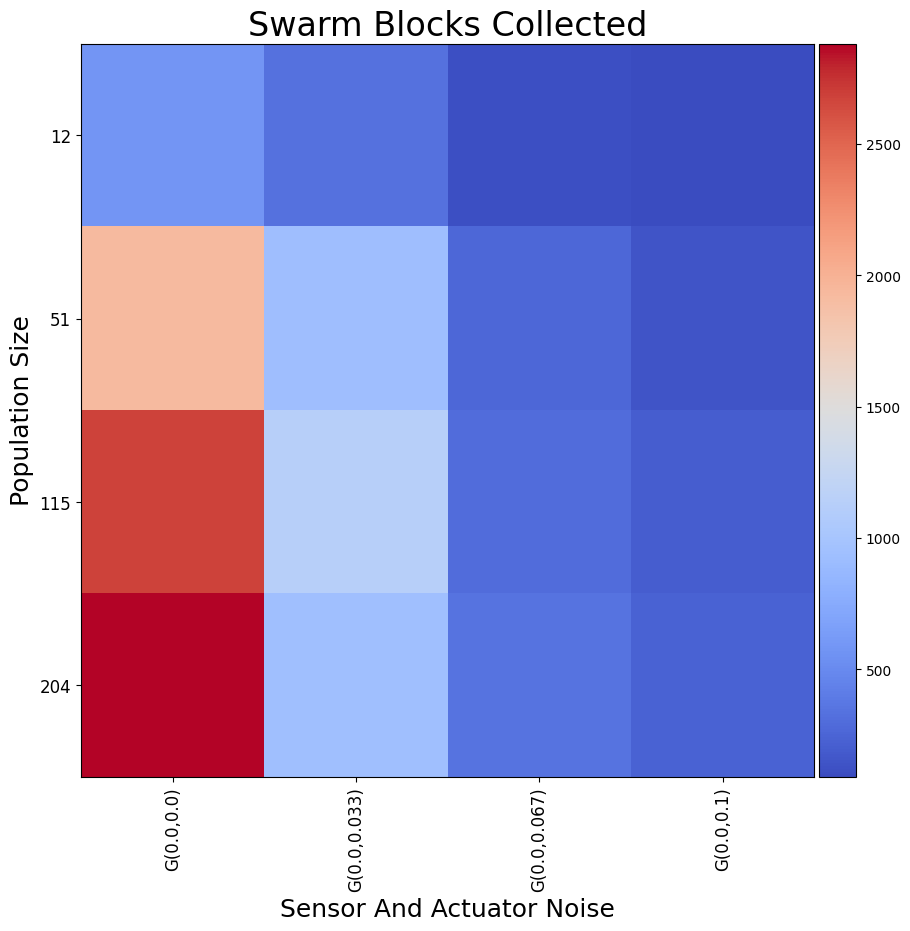}
    \caption{Demonstration of graphical deliverable generation in stage
      4. \emph{Left:} Intra-experiment heatmap showing the average locations
      where objects are found in the environment (nest in the
      center). \emph{Right:} Summary performance heatmap showing how the
      foraging behavior varies under a bivariate batch criteria with changing
      system size and different levels of sensor and actuator
      noise.\label{fig:stage4-ex-graphs}}
\end{figure}

Processed experimental results are used to generate deliverables to be part of
published research. This can include graphs or videos showing different aspects
of the system's response to the research query. Which graphs or videos are
generated is controlled by YAML configuration, allowing researchers to easily
disable generation of deliverables not of interest.  SIERRA's automation in this
stage makes it easy to modify a specific graph or video, if, for example, one
needs to be modified at a reviewer's request, eliminating the tedious process of
locating previously written throw-away scripts to regenerate it. For some
examples of the types of graphs SIERRA can generate during this stage,
see~\cref{fig:stage4-ex-graphs}.

One important feature of SIERRA in this stage is its model framework. It allows
researchers to generate data from first principles or from experimental results
(or both), and plot the generated data alongside empirical results; this is
commonly used for plotting model predictions. As with pipeline plugins, adding
new models is done by implementing a python interface, and placing the resulting
file(s) on SIERRA's plugin path. Models can be written in any language; only the
bindings must be written in python. An example of this capability is shown
in~\cref{fig:stage5-ex-graphs}.

In our first use case, suppose that Alice did not like the initial axes labels
on the heatmap generated in~\cref{fig:stage4-ex-graphs}. She could change her
YAML configuration files telling SIERRA what graphs she wanted to generate, and
then re-run SIERRA with the same command to regenerate the graph. In our second
use case, if Alice did not like the framerate of the rendered video she could
change the render command used to encode the video and regenerate it by adding
\verb|--render-cmd-opts --pipeline 4| to~\cref{fig:uc2:cmd}.

\subsection{Deliverable Comparison}
\begin{figure}[t]
  \centering
    \includegraphics[width=0.45\linewidth]{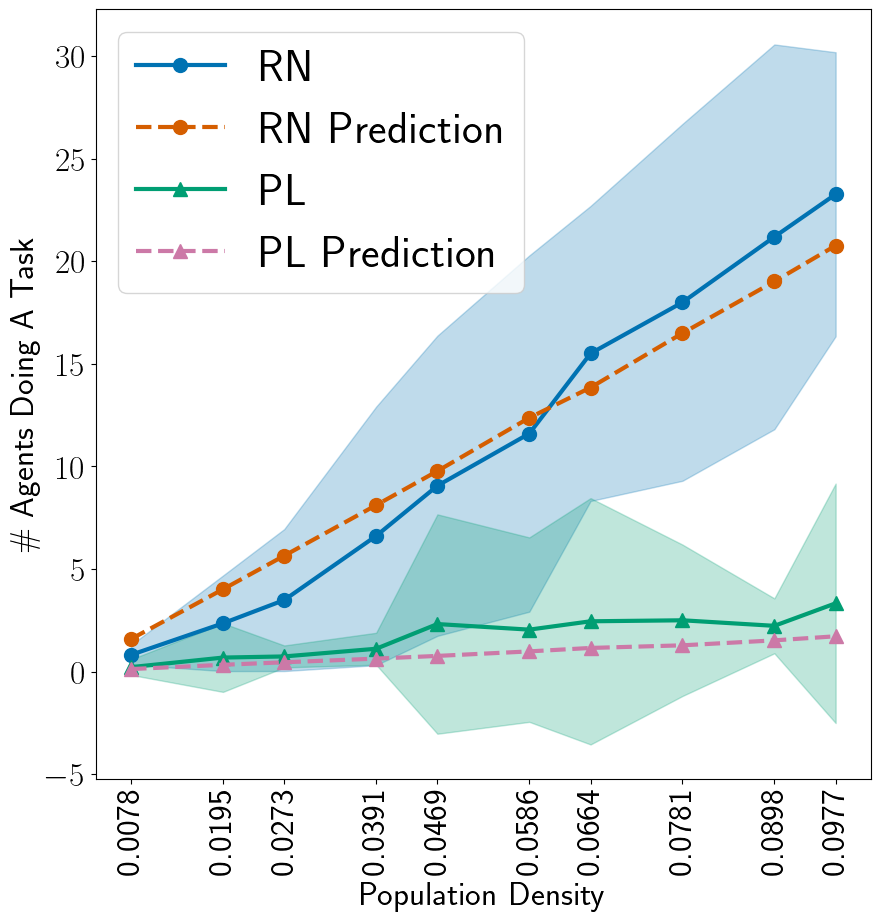}
    \includegraphics[width=0.54\linewidth]{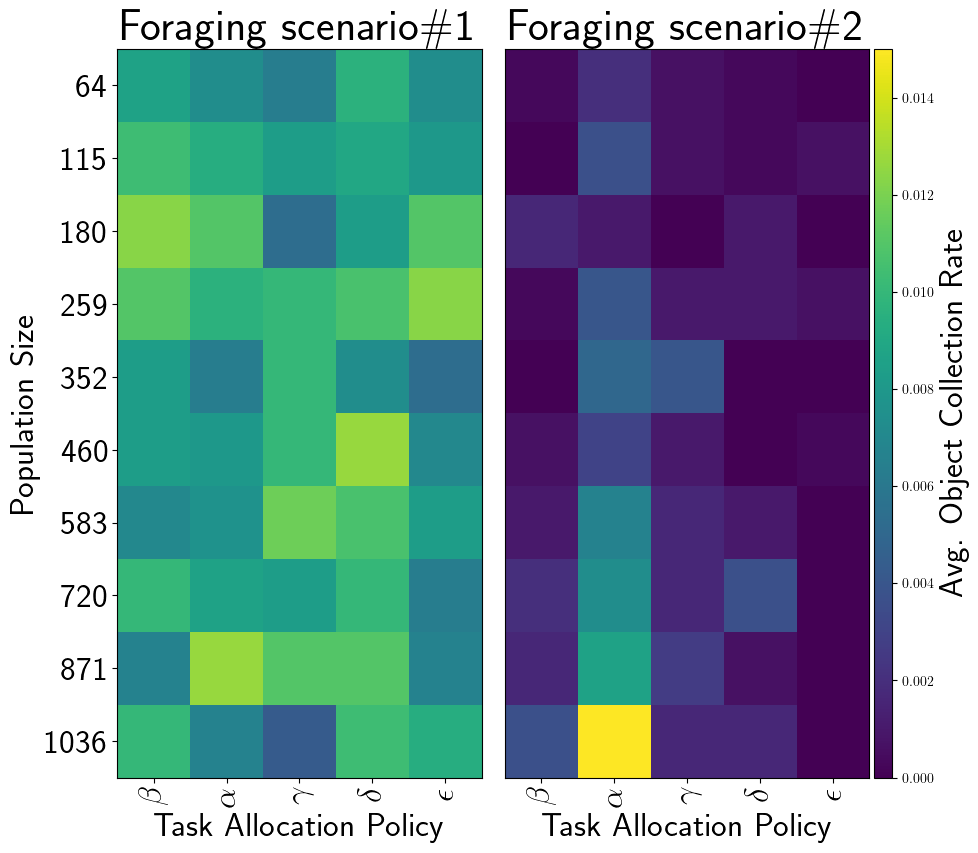}

    \caption{Demonstration of graphical deliverable comparison using both
      univariate and bivariate batch criteria. \emph{Left:} Intra-scenario
      deliverable comparison of several different task allocation algorithms on
      two different scenarios. \emph{Right:} Example inter-scenario comparison
      of a single algorithm ($\AlgA$). An analytical model was developed for
      $\AlgA$, and the resulting predictions plotted alongside actual
      data.\label{fig:stage5-ex-graphs}}
  \end{figure}

After deliverable generation, multiple deliverables can be combined to provide
side-by-side graphical comparisons; that is, SIERRA can take any data from two
graphs of any type from any two batch experiments and replot them on a single
figure. This comparison can take two forms.  First, \emph{intra-scenario}
comparison, in which graphs from experiments evaluating different algorithms in
the same context (scenario) are combined; this is shown
in~\cref{fig:stage5-ex-graphs}(a).  Second, \emph{inter-scenario} comparison, in
which graphs from batch experiments evaluating the same algorithm in different
contexts (scenarios) are combined; this is shown
in~\cref{fig:stage5-ex-graphs}(b).  Such high-level comparisons are useful for
demonstrating where/how a given AI element is better or different than the other
method; $\AlgA$ in our use cases. For example, suppose the Alice did not like
the side-by-side heatmaps showing differences in algorithm performance, because
they did not show which differences were statistically significant. She could
ask SIERRA to generate a set of linegraphs instead (one per row or column in the
heatmap), showing summary statistics such as confidence intervals or box and
whisker plots graphically.

\section{Discussion}\label{sec:discussion}
%
We have given a brief tour through some of SIERRA's features, using motivating
use cases to show how using SIERRA can address two of the most pressing needs in
intelligent systems research: increasing research reproducibility of results,
and accelerating research and development cycles. Furthermore, as illustrated by
our second use case, by addressing these two needs simultaneously SIERRA also
substantially lowers the bar to collaboration between researchers across
disciplines. Clearly, the scope of SIERRA's applicability is broad, and many
compelling use cases for its adoption exist. We believe that adoption of a tool
such as SIERRA to provide a near-universal pipeline for intelligent systems
research that supports reproducibility and reusability is paramount to
continuing to make meaningful progress as systems and approaches become more
complex.

However, SIERRA was originally developed for robotics research and therefore
with the needs of robotics researchers in mind, and its \emph{direct}
applicability to other domains may be limited, for two reasons.  First, as
described above, it is currently restricted to use cases where experimental
inputs can be specified in XML. This is not a major limitation in the robotics
community, as most of the major platforms support XML inputs. Outside robotics,
other platforms of interest might not support XML, making the utilization of
SIERRA impossible without substantial work on the part of researchers to develop
tools to translate non-XML input formats to/from XML so that SIERRA can work
with them. Nevertheless, if such translation \emph{is} necessary SIERRA's
modular design can easily support such an internal translation on a per-platform
basis for future use cases.

Second, the translation from automating agent-based or robotics research to
broader intelligent systems research involving machine learning, deep learning,
or other non-agent approaches may not fit the pipeline described
in~\cref{tab:research-pipeline}. For example, training runs for neural networks,
or qualitative coding of Human-Robot Interactions (HRI) studies may be difficult
to fit into SIERRA's paradigm. Nevertheless, for fields for which SIERRA may
be difficult to use, we would hope that it would serve as an inspirational basis
to build a similar tool.

Finally, we note two additional limitations of SIERRA.  First, SIERRA does not
attempt to homogenize configurations such that the results of a research query
are the same, regardless of platform and execution environment, which is not
possible in general. Second, SIERRA requires a common filesystem for all
automation components, which can include nodes in an HPC cluster, robots, etc.,
in order to be used out-of-the-box. Support for non-networked computational
components can of course be performed by researchers to adapt their code to
SIERRA, as shown in our second use case with a new \verb|--storage-medium|
plugin. Alternatively, if \verb|--no-master-node| is not passed then ROS
messages to pass robot data to a centralized collection node on the SIERRA host
machine to write out the data can be used--it would be as if the experiment had
run on the host machine from SIERRA's perspective.

\subsection{History and community acceptance}\label{ssec:discussion:hist}
SIERRA has been under development since 2016. Version \verb|1.2| was
demonstrated at AAMAS\cite{Harwell2022a-SIERRA}, and is available as a stable,
reliable release that comes with extensive documentation and tutorials. Earlier
versions of SIERRA have been used for several publications in top
conferences~\cite{Harwell2020a-demystify,Harwell2019a-metrics,Harwell2022a-ODE}
using ARGoS and ROS1 on PBS and SLURM HPC clusters, and with real TurtleBots at
the University of Minnesota. SIERRA is open source, and is available on
PyPI\footnote{\url{https://pypi.org/sierra-research}}. SIERRA receives about 100
downloads/week, a sign of its relevance as a tool for researchers.  A
comprehensive demonstration of SIERRA's capabilities can be found
here\footnote{\url{https://www-users.cse.umn.edu/~harwe006/showcase/aamas-2022-demo}},
including how it supports easy exploration of independent variables and
experimental data.

\section{Conclusions}\label{sec:conclusion}

We have presented SIERRA, a new tool that addresses two important needs in the
intelligent systems community: the need for better automation of engineering
tasks that many researchers perform and the need for better reproducibility of
research results.  As a ``low threshold, no ceiling'' tool, it significantly
lowers the barrier to collaboration between researchers across disciplines
without compromising customizability for advanced users. Thus, SIERRA is not
only relevant for the current needs of intelligent systems researchers, but also
for their future needs, and we strongly argue for its inclusion in any
researcher's toolbox.  Further improvements to SIERRA include: removing the
restriction that experimental inputs be specified in XML, expanding the set of
execution environments and platforms it supports natively, and refining its
configurability during statistics and graph generation to expose more of the
underlying \verb|matplotlib| and \verb|pandas|.




\bibliographystyle{IEEEtran}
\bibliography{bib/self,bib/platforms-and-tools,bib/reproducibility}

\begin{thebibliography}{10}
\providecommand{\url}[1]{#1}
\csname url@samestyle\endcsname
\providecommand{\newblock}{\relax}
\providecommand{\bibinfo}[2]{#2}
\providecommand{\BIBentrySTDinterwordspacing}{\spaceskip=0pt\relax}
\providecommand{\BIBentryALTinterwordstretchfactor}{4}
\providecommand{\BIBentryALTinterwordspacing}{\spaceskip=\fontdimen2\font plus
\BIBentryALTinterwordstretchfactor\fontdimen3\font minus
  \fontdimen4\font\relax}
\providecommand{\BIBforeignlanguage}[2]{{%
\expandafter\ifx\csname l@#1\endcsname\relax
\typeout{** WARNING: IEEEtran.bst: No hyphenation pattern has been}%
\typeout{** loaded for the language `#1'. Using the pattern for}%
\typeout{** the default language instead.}%
\else
\language=\csname l@#1\endcsname
\fi
#2}}
\providecommand{\BIBdecl}{\relax}
\BIBdecl

\bibitem{Afzal2021-reproduce}
A.~Afzal, D.~S. Katz, C.~Le~Goues, and C.~S. Timperley, ``Simulation for
  robotics test automation: Developer perspectives,'' in \emph{2021 14th IEEE
  Conference on Software Testing, Verification and Validation (ICST)}, 2021,
  pp. 263--274.

\bibitem{ROS2009}
M.~Quigley, K.~Conley, B.~Gerkey, J.~Faust, T.~Foote, J.~Leibs, R.~Wheeler, and
  A.~Ng, ``{ROS}: an open-source robot operating system,'' ICRA Workshop on
  Open Source Software, 01 2009.

\bibitem{SLURM2003}
A.~B. Yoo, M.~A. Jette, and M.~Grondona, ``{SLURM:} simple linux utility for
  resource management,'' in \emph{Job Scheduling Strategies for Parallel
  Processing}, D.~Feitelson, L.~Rudolph, and U.~Schwiegelshohn, Eds.\hskip 1em
  plus 0.5em minus 0.4em\relax Springer, 2003, pp. 44--60.

\bibitem{SwarmRob2018-reproduce}
A.~Pörtner, M.~Hoffmann, S.~Zug, and M.~Knig, ``{Swarmrob:} a docker-based
  toolkit for reproducibility and sharing of experimental artifacs in robotics
  research,'' in \emph{Proc. IEEE Int'l Conf. on Systems, Man, and Cybernetics
  (SMC)}, 2018, pp. 325--332.

\bibitem{Gundersen2019-reproduce}
O.~E. Gundersen, ``Standing on the feet of giants — reproducibility in ai,''
  \emph{AI Magazine}, vol.~40, pp. 9--23, 12 2019.

\bibitem{Gundersen2022-reproduce}
O.~E. Gundersen, S.~Shamsaliei, and R.~J. Isdahl, ``Do machine learning
  platforms provide out-of-the-box reproducibility?'' \emph{Future Generation
  Computer Systems}, vol. 126, pp. 34--47, 2022.

\bibitem{Gundersen2018-reproduce}
O.~E. Gundersen, Y.~Gil, and D.~W. Aha, ``On reproducible ai: Towards
  reproducible research, open science, and digital scholarship in ai
  publications,'' \emph{AI Magazine}, vol.~39, no.~3, pp. 56--68, Sep. 2018.

\bibitem{CodeOcean2019-reproduce}
A.~Clyburne-Sherin, X.~Fei, and S.~A. Green, ``Computational reproducibility
  via containers in social psychology,'' \emph{Meta-Psychology}, vol.~3, 2019.

\bibitem{DataFed2020-reproduce}
D.~Stansberry, S.~Somnath, G.~Shutt, and M.~Shankar, ``A systemic approach to
  facilitating reproducibility via federated, end-to-end data management,'' in
  \emph{Driving Scientific and Engineering Discoveries Through the Convergence
  of HPC, Big Data and AI}, J.~Nichols, B.~Verastegui, A.~Maccabe,
  O.~Hernandez, S.~Parete-Koon, and T.~Ahearn, Eds.\hskip 1em plus 0.5em minus
  0.4em\relax Springer Int'l Publishing, 2020, pp. 83--98.

\bibitem{WebotsHPC2021}
M.~Franchi, ``Webots.{HPC}: {A} parallel robotics simulation pipeline for
  autonomous vehicles on high performance computing,'' 2021.

\bibitem{Robotarium2016}
D.~Pickem, P.~Glotfelter, L.~Wang, M.~Mote, A.~Ames, E.~Feron, and
  M.~Egerstedt, ``The robotarium: A remotely accessible swarm robotics research
  testbed,'' 2016.

\bibitem{Bellogin2021-reproduce}
A.~Bellog{\'i}n and A.~Said, ``Improving accountability in recommender systems
  research through reproducibility,'' \emph{User Modeling and User-Adapted
  Interaction}, vol.~31, no.~5, pp. 941--977, Nov 2021.

\bibitem{NetLogo2004}
S.~Tisue, ``{NetLogo}: Design and implementation of a multi-agent modeling
  environment,'' in \emph{Proc. Agents 2004 Conference}, Oct. 2004.

\bibitem{ARGoS2012}
C.~Pinciroli \emph{et~al.}, ``{ARGoS}: a modular, parallel, multi-engine
  simulator for multi-robot systems,'' \emph{Swarm Intelligence}, vol.~6, pp.
  271--295, 12 2012.

\bibitem{Gazebo2004}
N.~Koenig and A.~Howard, ``Design and use paradigms for {G}azebo, an
  open-source multi-robot simulator,'' in \emph{2004 IEEE/RSJ International
  Conference on Intelligent Robots and Systems (IROS)}, vol.~3.\hskip 1em plus
  0.5em minus 0.4em\relax IEEE, 2004, pp. 2149--2154.

\bibitem{Webots2004}
O.~Michel, ``Webots: Professional mobile robot simulation,'' \emph{Journal of
  Advanced Robotics Systems}, vol.~1, no.~1, pp. 39--42, 2004.

\bibitem{TurtleBot2020}
R.~Amsters and P.~Slaets, ``Turtlebot 3 as a robotics education platform,'' in
  \emph{Robotics in Education}, M.~Merdan, W.~Lepuschitz, G.~Koppensteiner,
  R.~Balogh, and D.~Obdr{\v{z}}{\'a}lek, Eds.\hskip 1em plus 0.5em minus
  0.4em\relax Springer International Publishing, 2020, pp. 170--181.

\bibitem{GNUParallel2011}
\BIBentryALTinterwordspacing
O.~Tange, ``Gnu parallel - the command-line power tool,'' \emph{;login: The
  USENIX Magazine}, vol.~36, no.~1, pp. 42--47, Feb. 2011. [Online]. Available:
  \url{http://www.gnu.org/s/parallel}
\BIBentrySTDinterwordspacing

\bibitem{Harwell2022a-SIERRA}
J.~Harwell, L.~Lowmanstone, and M.~Gini, ``Sierra: A modular framework for
  research automation,'' in \emph{Proc. Int'l Conf. on Autonomous Agents and
  Multiagent Systems (AAMAS)}, 2022, p. 1905–1907.

\bibitem{Harwell2020a-demystify}
------, ``Demystifying emergent intelligence and its effect on performance in
  large robot swarms,'' in \emph{Proc. Int'l Conf. on Autonomous Agents and
  Multi-Agent Systems (AAMAS)}, May 2020, pp. 474--482.

\bibitem{Harwell2019a-metrics}
J.~Harwell and M.~Gini, ``Swarm engineering through quantitative measurement of
  swarm robotic principles in a 10,000 robot swarm,'' in \emph{Proc. 28th Int'l
  Joint Conf. on Artificial Intelligence (IJCAI-19)}, Aug. 2019, pp. 336--342.

\bibitem{Harwell2022a-ODE}
J.~Harwell, A.~Sylvester, and M.~Gini, ``Characterizing the limits of linear
  modeling of non-linear swarm behaviors,'' arXiv:2110.12307v2 [cs.RO], 2022.

\end{thebibliography}

\end{document}